\documentclass[11pt]{article}

\usepackage[final]{acl}

\usepackage{times}
\usepackage{latexsym}

\usepackage[T1]{fontenc}

\usepackage[utf8]{inputenc}

\usepackage{microtype}

\usepackage{inconsolata}

\usepackage{graphicx}
\usepackage{amsmath} 
\usepackage{amsfonts}
\usepackage{booktabs} 
\usepackage{multirow} 
\usepackage{enumitem}
\usepackage{makecell}
\usepackage{dsfont} 
\usepackage{amssymb}
\usepackage{array}
\usepackage{tabularx} 

\title{Why LLMs Hallucinate on Structured Knowledge: A Mechanistic Analysis of Reasoning over Linearized Representations}

\author{
\textbf{Shanghao Li}$^{1}$, \textbf{Jinda Han}$^{2}$, \textbf{Yibo Wang}$^{1}$, \textbf{Yuanjie Zhu}$^{1}$,\\
\textbf{Zihe Song}$^{1}$, \textbf{Langzhou He}$^{1}$, \textbf{Kenan Kamel A Alghythee}$^{1}$, \textbf{Philip S. Yu}$^{1}$ \\
$^{1}$University of Illinois Chicago \\
$^{2}$University of Illinois Urbana-Champaign \\
\texttt{\{sli261, ywang633, yzhu224, zsong29, lhe24, kalghy2, psyu\}@uic.edu} \\
\texttt{jhan51@illinois.edu}
}

\begin{document}
\maketitle

\begin{abstract}
In many reasoning tasks, large language models (LLMs) rely on structured external knowledge, such as graphs and tables, which is typically linearized into sequential token representations.
However, even when sufficient knowledge is available, LLMs can still produce hallucinated outputs, and the underlying mechanisms behind such failures remain poorly understood.
We investigate these mechanisms and find that hallucinations arise from systematic internal dynamics rather than random noise.
First, attention disproportionately concentrates toward shortcut-like structural cues rather than distributing across the full context. 
Second, feed-forward representations fail to ground the provided knowledge, causing the model to revert to parametric memory.
Moreover, our results indicate that hallucination is consistently associated with failures in semantic grounding within feed-forward layers, while attention allocation exhibits greater task-dependent variability.
Finally, we show that these mechanistic patterns generalize beyond single-hop graphs to multi-hop and tabular settings, enabling effective hallucination detection across structured knowledge formats.

\end{abstract}
\vspace{-0.15in}

\section{Introduction} 

In real world, a vast portion of human knowledge is stored in structured formats, such as knowledge graphs and tables. 
When using such structured knowledge for reasoning tasks with LLMs, a common practice is to linearize it into textual sequences \cite{jin2024large, deng2024graphvis, deng-etal-2024-tables}, as current LLM architectures operate exclusively over sequential token representations.
For instance, modern Retrieval-Augmented Generation (RAG) frameworks typically serialize retrieved subgraphs into relational triplets or convert tabular data into textual descriptions to facilitate downstream reasoning \cite{li2025subgraphrag, kim-etal-2023-kg, chen2024tablerag, zhang2025aixelask}.

Despite sufficient and accurate knowledge provided in the input, LLMs frequently produce hallucinated outputs when reasoning over linearized structured knowledge \cite{ming2024faitheval, sun2024redeep}.
While existing literature primarily addresses this through external interventions such as retrieval augmentation and prompt engineering at the input or output level \cite{huang2025survey, agrawal2024can}, a significant gap remains in understanding the underlying mechanistic drivers: 
\textit{what causes models to underutilize explicit structured knowledge already present in their input, leading to hallucinated responses?}

We hypothesize that these hallucinations are not random noise, but rather the result of systemic internal failures arising from the inherent tension between linearized structures and the Transformer’s inductive bias toward modeling natural language as sequential text. 
To move beyond black-box observations, we examine two core functional components of Transformer models: attention heads, which selectively attend to subsets of the input \cite{michel2019sixteen, voita-etal-2019-analyzing}, and feed-forward networks (FFNs), which play a central role in storing and integrating knowledge \cite{dai2022knowledge, geva2021transformer}. 
By probing model behavior through these two complementary lenses, we investigate how attention heads and FFNs collaboratively linearized representations of structured knowledge, and how dysfunctions in these processes lead to the emergence of hallucinations.

\begin{figure*}[t]
   \centering
   \includegraphics[width=\linewidth]{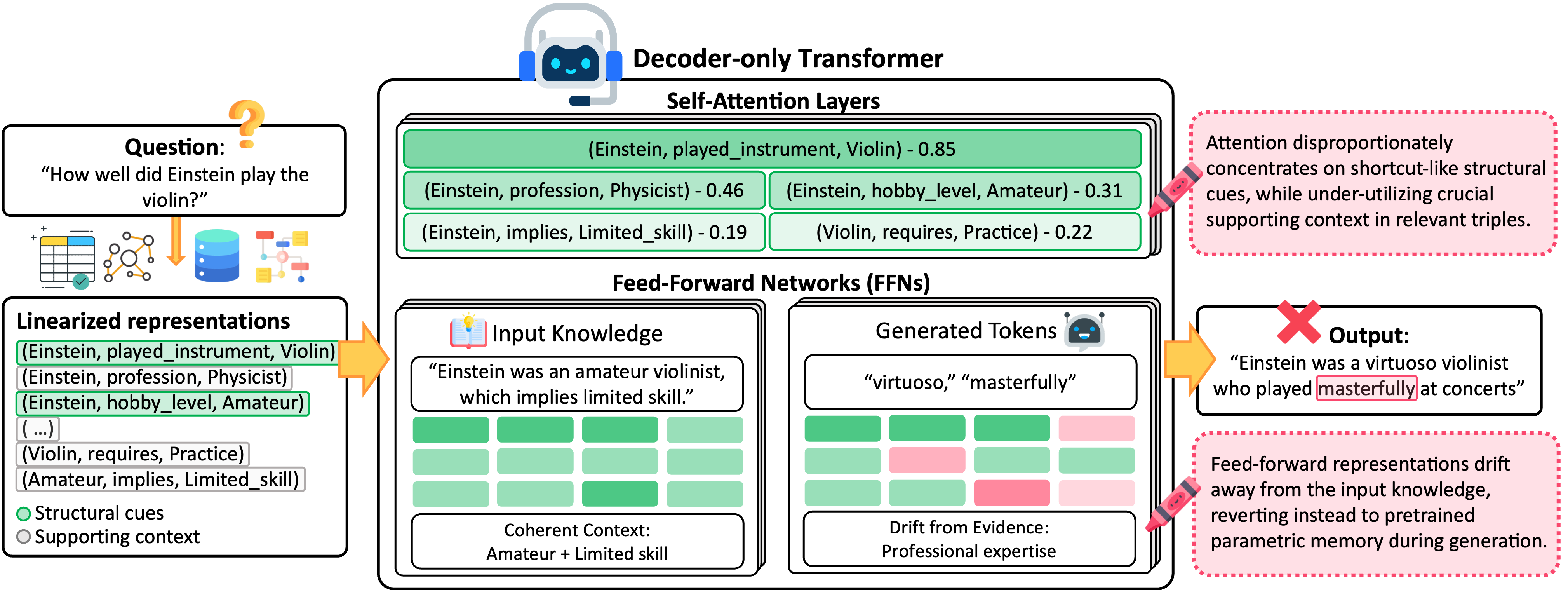}
   \vspace{-0.25in}
    \caption{An illustrative overview of hallucination mechanisms in structured knowledge reasoning. The figure highlights two recurring internal failure patterns when linearized structured knowledge is processed by Transformer models: over-reliance on shortcut-like structural cues in attention, and semantic misalignment in feed-forward representations that leads to drift from retrieved evidence.}
   \label{fig: hallucinations-reasons}
   \vspace{-.15in}
\end{figure*}

Building on prior findings that linearization introduces structural redundancy and that LLMs are prone to shortcut learning, we posit that hallucinations arise from a systematic imbalance between how Transformer models select structural cues and how they integrate semantic evidence (As shown in Figure~\ref{fig: hallucinations-reasons}.)
To investigate this hypothesis, we leverage mechanistic interpretability to decouple the model's utilization of external evidence and internal memory.
We introduce two metrics: \textbf{Structural Shortcut Reliance (SSR)}, which uses attention heads to quantify the concentration of attention on structural regularities in linearized inputs, and \textbf{Semantic Alignment Score (SAS)}, which employs (FFNs) to evaluate the alignment between internal representations and supporting evidence during generation.

Across evaluations on single-hop and multi-hop graph reasoning as well as tabular data, our correlation and distributional analyses reveal two recurrent patterns: 
(i) Attention over-concentrates on shortcut-like structural cues (e.g., minimal query–answer paths in linearized knowledge graph triplets)
(ii) feed-forward representations fail to maintain semantic grounding in the provided knowledge, with internal states drifting away from the evidence and reverting to parametric memory. 
Moreover, a quadrant-based analysis shows that attention patterns, captured by SSR, vary with task complexity and the format of structured knowledge, whereas the ability to maintain semantic grounding in internal representations, captured by SAS, plays a more stable role in mitigating hallucination.
Importantly, these mechanistic signals are not merely descriptive—they enable practical intervention. We demonstrate that SSR and SAS can be used to build a lightweight, plug-and-play hallucination detector that outperforms existing confidence-based and consistency-based baselines, without any model modification or fine-tuning.

\section{Background and Related Work}
Prior work on LLM interpretability has investigated how internal mechanisms such as self-attention and hidden representations influence model behavior. 

\subsection{Attention Heads} 
Although attention weights are not faithful explanations of model decisions, they provide useful insight into how Transformers route information and allocate focus across a sequence \cite{jain2019attention, serrano2019attention, wiegreffe2019attention}. 
Importantly, this allocation is highly selective rather than uniform: many attention heads consistently focus on a limited subset of tokens instead of integrating the entire context. 
Prior analyses, for instance, have shown attention heads that repeatedly attend to specific linguistic or structural elements, such as syntactic relations or special markers \cite{clark-etal-2019-bert}. 

When structured knowledge is linearized into a sequence, we expect this selectivity to pose significant challenges because the original relational dependencies are no longer explicitly represented. 
As a result, attention allocation tends to concentrate on a small portion of the input, potentially leaving other crucial revelant evidence underused and resulting in topological coverage failure. 
In this work, we examine whether and how such attention concentration correlates with hallucination when LLMs reasoning over linearized structured knowledge.

\subsection{Feed-forward layers (FFNs)}
While self-attention aggregates token-level information, feed-forward networks (FFNs) transform these signals into intermediate representations that directly shape model outputs.
As the primary substrate for storing parametric knowledge \cite{geva2021transformer}, FFNs do not merely pass through selected information; they actively modify it based on patterns learned during pretraining \cite{geva2022transformer, kobayashi2024analyzing}. 
Prior work suggests that when external evidence is weak or ambiguous, internal parametric knowledge can override the retrieved context, leading to knowledge-driven hallucinations rooted in the model's priors \cite{tao-etal-2024-context, farahani-johansson-2024-deciphering}.

A critical vulnerability arises when processing structured knowledge, which must be linearized into token sequences. Unlike natural language, whose syntactic structure implicitly preserves relational meaning, linearization removes explicit structural constraints. This flattening weakens the semantic scaffolding for FFNs, potentially allowing parametric priors to dominate and induce representation drift. 
In this work, we investigate whether and how the alignment between internal model representations and linearized structured knowledge correlates with hallucination.

\paragraph{Synthesis.}
In summary, prior mechanistic studies have deepened our understanding of attention and representation in Transformers, largely in the context of unstructured natural language. Our work addresses a critical gap by examining how attention and feed-forward representations interact and break down when processing structured knowledge that has been flattened into token sequences.

\section{Mechanistic Diagnostic Metrics}
We introduce two diagnostic metrics—Structural Shortcut Reliance (SSR) and Semantic Alignment Score (SAS)—to probe the internal mechanisms of structured-knowledge reasoning.
Their implementation details and interpretations are provided in Appendix~\ref{appendix:pruning} and Appendix~\ref{sec:metric_interpretation}, respectively.
We further verify that both metrics capture meaningful signals rather than artifacts of structural heuristics or superficial lexical overlap, through controlled analyses (Appendix~\ref{sec: linearization order}, Appendix~\ref{sec: support set}).

\subsection{Structural Shortcut Reliance (SSR)}
We posit that attention allocation becomes increasingly concentrated on a limited subset of structural cues when large language models process structured knowledge that has been flattened into token sequences. In particular, the model may concentrate attention on a minimal set of structurally salient cues (which we later refer to as \emph{core structural cues}) that appear to directly connect the query to a plausible answer, instead of integrating the full set of supporting evidence.
To quantify this attention bias in structured knowledge settings, we propose \emph{Structural Shortcut Reliance (SSR)}, which measures \emph{the extent to which a model’s attention over-relies on a minimal set of shortcut-like structural cues during answer generation}.

\paragraph{Core structural cues.}
We define \emph{core structural cues} ($S$) as the minimal token subset establishing a direct connection between query and answer. While these cues may trigger a candidate answer through learned associations, but often lack the comprehensive relational constraints required for factual verification.
The complement set $\bar{S}$ contextual cues provides the relational environment and global constraints necessary to ground and validate the direct connection.
We posit that an over-reliance on $S$ at the expense of $\bar{S}$ indicates that the model is bypassing contextual verification, leading to hallucination. 

\paragraph{Formulation.}
Let $A=\{a_1,\dots,a_n\}$ denote the positions of the generated answer tokens. For each decoder layer $l\in[1,L]$ and attention head $h\in[1,H]$, let $e_{l,h,i,j}$ be the raw attention score from answer position $i \in A$ to source position $j$. We apply the softmax function to obtain the normalized attention weights:
\begin{equation}
\alpha_{l,h,i,j} = \frac{\exp(e_{l,h,i,j})}{\sum_{k} \exp(e_{l,h,i,k})}.
\end{equation}
The total attention mass allocated to the core structural cues $S$ and the contextual structural cues $\bar{S}$ is defined as:
\begin{equation}
\alpha_{l,h,i,S} = \sum_{j \in S} \alpha_{l,h,i,j}, \quad \alpha_{l,h,i,\bar{S}} = \sum_{j \in \bar{S}} \alpha_{l,h,i,j}.
\end{equation}
We define the SSR as the average difference between the attention mass flowing to core cues versus contextual cues:
\begin{equation}
\text{SSR} = \frac{1}{L \cdot H \cdot |A|} \sum_{l=1}^{L} \sum_{h=1}^{H} \sum_{i \in A} \bigl( \alpha_{l,h,i,S} - \alpha_{l,h,i,\bar{S}} \bigr).
\end{equation}
Since $\alpha_{l,h,i,S} + \alpha_{l,h,i,\bar{S}} = 1$ (assuming $S$ and $\bar{S}$ partition the input sequence), SSR is bounded within $[-1, 1]$.

\paragraph{Instantiation.}
In our study, we instantiate these sets across different knowledge formats. For graph-based data, $S$ corresponds to the shortest path triples.
For tabular data, $S$ refers to the core structural triples (typically 1-2) that are extracted from specific rows and simultaneously contain both the question entities (or match the question column headers) and the answer cells, thereby directly fulfilling the query's relational constraints.

\subsection{Semantic Alignment Score (SAS)}
\label{sec:SAS}
Prior work has shown that feed-forward layers (FFNs) may fail to faithfully integrate retrieved evidence, instead relying on pretrained parametric knowledge \cite{sun2024redeep, tao-etal-2024-context}. 
We argue that this tendency is exacerbated when structured knowledge is linearized into token sequences, since even when attention is properly allocated, semantic content is fragmented during linearization, weakening the constraints available for representation fusion within FFNs.
To quantify this representation-level drift, we propose the \emph{Semantic Alignment Score (SAS)}, which measures the extent to which a model’s internal answer representations remain semantically grounded in the structured knowledge provided as input during generation.

\paragraph{Supporting Context Set (SCS).}
To provide a semantic reference for evaluating whether a model’s internal representations remain grounded during generation, we define the \emph{Supporting Context Set} ($\mathcal{E}$) as a set of input elements that collectively capture the semantic context relevant to the reasoning process. 
We build SCS by expanding from the core structural cues ($S$), incorporating their immediate structural neighbors so as to include sufficient relational context while remaining closely tied to the underlying reasoning structure.

\paragraph{Instantiation.}
We instantiate $\mathcal{E}$ based on the structural characteristics of the input. 
For graph-based data, $\mathcal{E}$ expands the core structural cues ($S$) by including their one-hop neighboring triples to capture local relational density. 
For tabular data, $\mathcal{E}$ expands $S$ by including semantically relevant triples that contain question entities, answer cells, or match question column names, providing the necessary contextual environment for the target cells.

\paragraph{Formulation.}
For each answer token $y_t$, we extract its hidden representation $\mathbf{h}_t$ from a penultimate Transformer layer before the final output projection. Each knowledge unit $U_i \in \mathcal{E}$ (e.g., a triple or a linearized table row) is encoded by the same language model to obtain a reference embedding $\mathbf{g}_i$. After $\ell_2$ normalization, the token-level alignment is defined as:
\begin{equation}
 \text{SAS}(y_t) = \max_{U_i \in \mathcal{E}} \cos\left( \mathbf{h}_t,\, \mathbf{g}_i \right)
\end{equation}
The overall sentence-level SAS is the average across all $T$ answer tokens:
\begin{equation}
    \text{SAS} = \frac{1}{T}\sum_{t=1}^{T}\text{SAS}(y_t).
\end{equation}
SAS is theoretically bounded in $[-1, 1]$, where values closer to 1 indicate that the model's internal representation is semantically grounded in the provided evidence. 

\section{Mechanistic Experiment Setup}
This section describes the experimental setup used to analyze how the proposed mechanistic metrics relate to hallucination behavior under controlled reasoning settings.

\subsection{Hypotheses and Overview}
Our analysis is guided by three hypotheses regarding the representation-level mechanics of structured knowledge processing:


\begin{itemize}
\vspace{-0.2cm}
    \item \textbf{H1 (Distributional Divergence).}
    Hallucinated and non-hallucinated generations exhibit systematically divergent distributions of SSR and SAS; specifically, hallucinated outputs tend to exhibit greater reliance on core structural cues (higher SSR) and weaker semantic grounding (lower SAS). 
    \vspace{-0.2cm}
    \item \textbf{H2 (Complementarity).}
    SSR and SAS capture complementary aspects of the model’s internal processing during hallucination, such that combining the two signals yields a more informative account of hallucination than either metric alone. 
    \vspace{-0.2cm}
    \item \textbf{H3 (Joint Failure Modes).}
    Specific joint configurations of SSR and SAS correspond to mechanistically distinct hallucination patterns within the model’s reasoning process.
\end{itemize}

To test these hypotheses, we feed linearized knowledge graph (KG) subgraphs into a frozen LLM using greedy decoding for deterministic generation.
We explicitly define core structural cues ($S$) and supporting context ($\mathcal{E}$) within the input sequences to measure attention toward $S$ (SSR) and the semantic alignment of hidden states with $\mathcal{E}$ (SAS) during reasoning. 
Generated responses are categorized by factual correctness; we then statistically compare the SSR and SAS distributions across hallucinated and non-hallucinated groups. 
Finally, we evaluate the generalizability of these metrics across multi-hop reasoning and tabular data (Section~\ref{sec: generalization}).

\subsection{Dataset and Experimental Setting}
\label{sec:setup}
\paragraph{Dataset and Model}
We analyze 5{,}000 randomly sampled examples from the MetaQA-1hop development set \cite{zhang2018variational}, a knowledge graph question-answering benchmark in the movie domain.
Following established practice in mechanistic analysis \cite{sun2024redeep, wang2025seredeep}, we provide the model with oracle subgraphs constructed in prior work \cite{He-WSDM-2021, sun2018open} to decouple knowledge processing from retrieval noise.
For this mechanistic analysis, we use \texttt{Llama-2-7b-chat-hf} \cite{touvron2023llama} as the base generator.

\paragraph{Subgraph Linearization and Prompting.}
For each question, We first identify the \emph{Core Structural Cues} ($S$) by extracting all triples on the shortest paths between question and gold answer entities. 
To construct the \emph{Supporting Context Set} ($\mathcal{E}$), we expand $S$ by incorporating its structural neighbors, ranked by their connectivity scores to key entities, until a total of $K=20$ triples is reached. 
This procedure ensures that the input contains both the essential reasoning backbone ($S$) and sufficient relational density ($\mathcal{E}$) required for stable semantic alignment analysis. 
To bridge the structural data with the LLM's natural language interface, we serialize the graph into subject-predicate-object triples (e.g., ``Conspiracy release\_year 2001'')  and integrate them into a fixed prompt template. 

\begin{figure}[t]
   \centering
   \vspace{-.15in}
   \includegraphics[width=\linewidth]{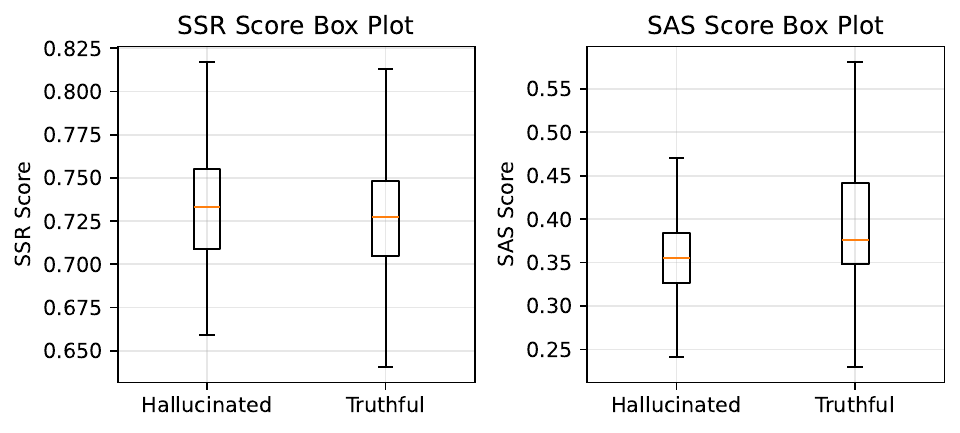}
   \caption{Box plots of SSR and SAS scores for hallucinated vs. truthful responses. Hallucinated answers show higher SSR and lower SAS, indicating increased reliance on structural cues and weaker semantic integration of the supporting context.}
   \label{fig:boxplots}
   \vspace{-.15in}
\end{figure}

\paragraph{Hallucination Labeling and Evaluation}
\label{sec:lallucination_label}
We label model outputs as hallucinated or non-hallucinated using SQuAD-style answer matching \cite{seo2016bidirectional}.
Specifically, we normalize generated answers through standard text cleaning and evaluate them against gold answers using Exact Match (EM) and token-level F1.
A response is labeled as non-hallucinated if it achieves either EM or an F1 score above a threshold ($F1 \geq 0.3$); otherwise, it is labeled as hallucinated.




\section{Mechanistic Findings and Analysis}
\subsection{Distributional Divergence and Statistical Significance} 
To test H1 (Distributional Divergence), we evaluate the discriminative power of the proposed metrics by comparing their distributions across hallucinated and non-hallucinated outputs.

As shown in Fig.~\ref{fig:boxplots}, SSR and SAS exhibit clear distributional differences between hallucinated and non-hallucinated outputs.
Two-sample t-tests confirm that these differences are statistically significant, with $t(4998) = -3.31$ ($p < 0.001$) for SSR and $t(4998) = 10.96$ ($p < 10^{-26}$) for SAS.
In particular, hallucinated outputs are associated with higher SSR and lower SAS scores.
These results support \textbf{H1}, indicating that hallucinated and non-hallucinated generations exhibit systematically divergent distributions of SSR and SAS. 

This further indicates that hallucinations in reasoning over linearized structured knowledge are not random errors, but are systematically associated with the model’s internal processing dynamics.
In particular, hallucinated responses correspond to states where attention disproportionately concentrates on shortcut-like structural cores (e.g., paths linking the query to the answer in the subgraph), while the model’s latent representations remain weakly grounded in the supporting context.

\subsection{Joint Feature Behavior and Complementarity}
\begin{figure}[h]
   \centering
   \setlength{\abovecaptionskip}{2pt} 
   \setlength{\belowcaptionskip}{-3pt} 
   \includegraphics[width=\linewidth]{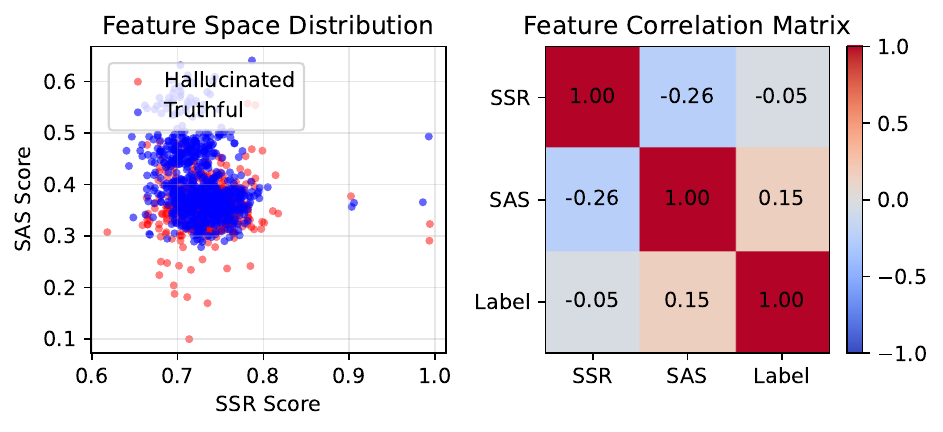}
   \caption{ Feature space and correlation. (\textit{Left}) Scatter plot of SSR vs. SAS shows overlapping but distinguishable clusters. (\textit{Right}) Correlation matrix shows weak dependencies between features and label, suggesting complementary information.}
   \label{fig:feature_analysis}
\end{figure}

To test H2 (Complementarity), we examine the joint distribution and Pearson correlation between SSR and SAS (Fig.~\ref{fig:feature_analysis}). 
The two signals exhibit modest negative correlation ($r = -0.26$).
Their individual correlations with hallucination labels are relatively low ($r_{SSR} = -0.05, r_{SAS} = 0.15$)
Together, these observations provide direct empirical support for \textbf{H2}, indicating that SSR and SAS capture distinct and decoupled aspects of the model’s internal behavior during reasoning using linearized structured knowledge.

Together, these results indicate that SSR and SAS capture distinct and only weakly coupled aspects of the model’s internal behavior during reasoning over linearized structured knowledge. 
Although increases in shortcut-oriented attention tend to co-occur with reduced semantic grounding, the modest correlation confirms that the two signals reflect complementary failure mechanisms rather than redundant indicators.

\vspace{-0.5\baselineskip} 

\subsection{Quadrant-Based Case Study of Hallucination Patterns}
\begin{figure}[h]
\centering
\setlength{\abovecaptionskip}{2pt} 
\setlength{\belowcaptionskip}{-3pt} 
\includegraphics[width=1.0\linewidth]{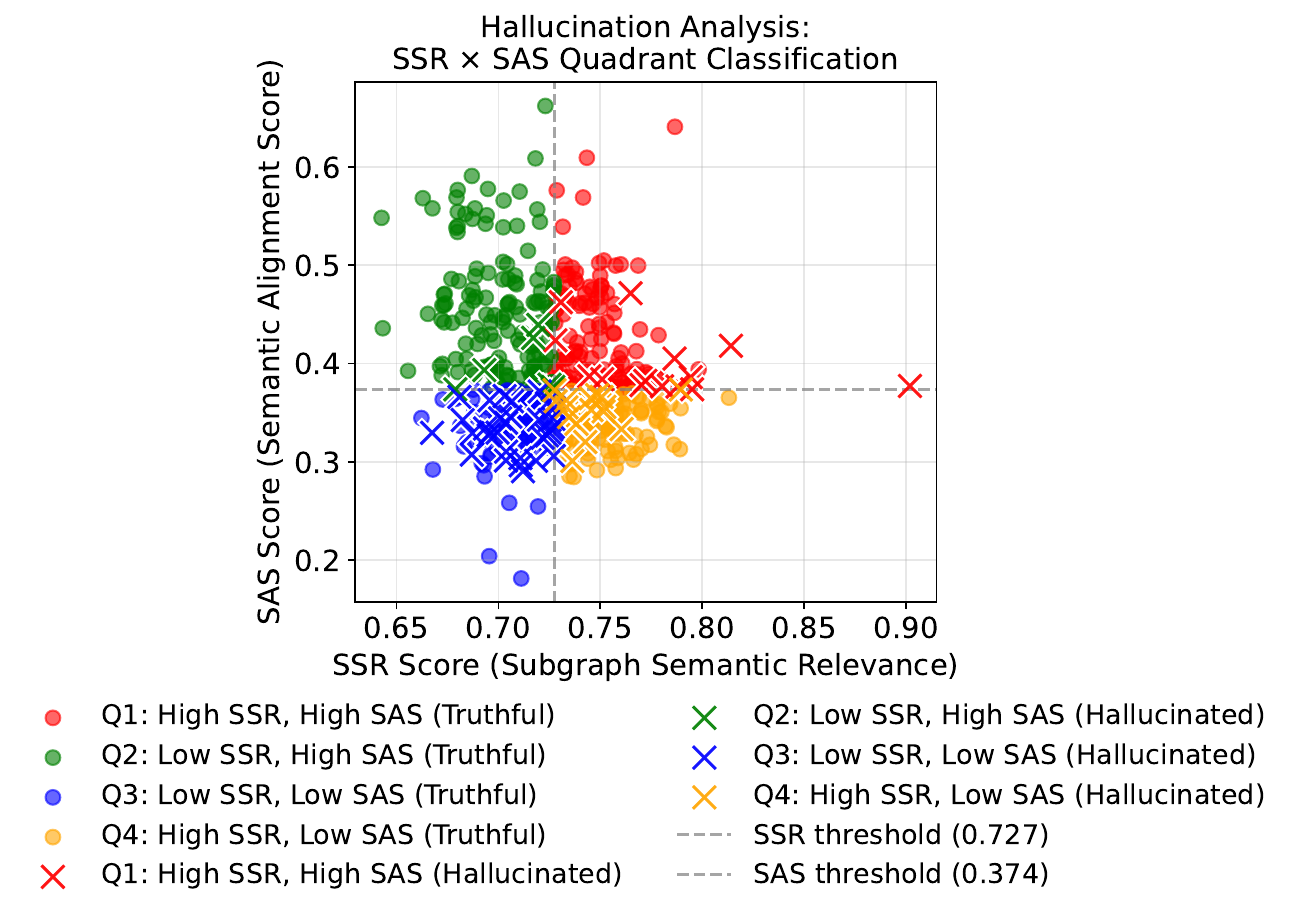}
\caption{Hallucination Analysis via SSR × SAS Quadrant Classification. Each point represents a model output. Hallucinated answers are marked with ``×''.}
\label{fig:quadrant-hallucination}
\end{figure}

To test H3 (Joint Failure Modes), we examine the interaction between attention focus and semantic alignment by partitioning model outputs into four quadrants based on median SSR ($0.727$) and SAS ($0.374$) values (Fig.~\ref{fig:quadrant-hallucination}):

\begin{itemize}
\vspace{-0.2cm}
\item \textbf{Q1 (High SSR, High SAS)}:
The model concentrates attention on Core Structural Cues ($S$) while successfully integrating the Supporting Context Set ($\mathcal{E}$). 
This configuration reflects a tendency for minimal reasoning paths that remain semantically aligned with the broader context.
\vspace{-0.2cm}
\item \textbf{Q2 (Low SSR, High SAS)}: 
The model allocates attention broadly beyond $S$ while effectively integrating information from $\mathcal{E}$ into its internal representations. 
This configuration corresponds to robust semantic grounding driven by comprehensive context integration.
\vspace{-0.2cm}
\item \textbf{Q3 (Low SSR, Low SAS)}: 
The model distributes attention broadly across both $S$ and $\mathcal{E}$ but fails to semantically integrate this information. 
This configuration reflects limited semantic integration of the attended information into coherent internal representations.
\vspace{-0.2cm}
\item \textbf{Q4 (High SSR, Low SAS)}: 
The model concentrates attention on $S$ while failing to effectively integrate information from $\mathcal{E}$. 
This configuration reflects reliance on minimal reasoning paths without sufficient semantic support from the broader context.
\end{itemize}

As summarized in Table~\ref{tab:quadrant-stats}, hallucination risk varies substantially across joint SSR–SAS configurations.
The lowest hallucination rate is observed in Q2 (Low SSR, High SAS; 5.0\%), where broad attention is accompanied by strong semantic alignment.
In contrast, risk increases sharply in configurations with low semantic alignment, peaking in Q3 (Low SSR, Low SAS; 22.2\%).
Notably, Q4 (High SSR, Low SAS) exhibits elevated but lower risk (10.9\%) than Q3, suggesting that weakened semantic grounding is a necessary but not sufficient condition for severe hallucination behavior.
These distinct risk profiles directly support \textbf{H3}, confirming that specific joint configurations of SSR and SAS correspond to mechanistically different processing states.

\vspace{-0.8\baselineskip} 

\begin{table}[h]
\centering
\setlength{\abovecaptionskip}{1pt} 
\setlength{\belowcaptionskip}{-1pt} 
\caption{Quadrant-level hallucination rate and average SSR/SAS values. Q3 (low focus, low grounding) shows the highest hallucination rate.}
\small
\begin{tabular}{lcccc}
\toprule
\textbf{Quadrant} & \textbf{Hall. (\%)} & \textbf{SSR} & \textbf{SAS} \\
\midrule
Q1: High SSR, High SAS & 9.5\%  & 0.752 & 0.421 \\
Q2: Low SSR, High SAS  & \textbf{5.0\%}  & 0.701 & 0.452 \\
Q3: Low SSR, Low SAS   & 22.2\% & 0.707 & 0.340 \\
Q4: High SSR, Low SAS  & 10.9\% & 0.754 & 0.344 \\
\bottomrule
\end{tabular}
\label{tab:quadrant-stats}
\end{table}

While individual metrics capture partial signals, the quadrant analysis reveals that hallucination cannot be explained by either SSR or SAS alone. 
Instead, distinct failure regimes emerge from their interaction, reflecting how structural attention routing and semantic grounding jointly shape model behavior.

To provide additional evidence for the robustness of our findings, we conduct additional analyses across model scales, reasoning depths, and input variations, including newer models (Qwen3), multi-hop benchmarks (MetaQA-3hop and ComplexWebQuestions (CWQ)~\cite{talmor-berant-2018-web}), and controlled perturbations of the input structure. 
Across all settings, the SSR--SAS relationship and quadrant structure remain consistent. Detailed results are provided in Appendix~\ref{sec:robustness}.

\section{Generalization and Practical Application}

\subsection{Generalization: Beyond Simple Graphs to Tables and Multi-hop Reasoning}
\label{sec: generalization}
To examine whether our mechanistic findings extend beyond simple graph-based reasoning, we evaluate the SSR--SAS framework under two more challenging settings: multi-hop graph reasoning and table-based question answering. 
These settings involve multi-step reasoning and alternative structured representations, enabling us to evaluate the generalizability of our primary findings.

\paragraph{Methodology.}

First, to examine how our findings extend to multi-step reasoning, we conduct an analysis on the development set of the \texttt{MetaQA-2hop} dataset, where the core structural cues ($\mathcal{S}$) and supporting context set ($\mathcal{E}$) are constructed in the same manner as in the 1-hop setting.

Second, to examine how our findings extend to other forms of structured representations, we conduct an analysis on table-based question answering using $2{,}000$ samples from the \texttt{WikiTableQuestions} dataset \cite{pasupat-liang-2015-compositional}. We preserve methodological consistency by  linearizing each table row into a set of triple-like units of the form 
$(\text{Entity}, \text{Column}, \text{Value})$, where each cell is represented 
as a separate triple linking the row entity to a column header and its value, following common table linearization strategies used in LLM-based reasoning \cite{min-etal-2024-exploring, sui2024table}. 
Under this representation, the core structural cues ($S$) correspond to column–cell associations that connect question-relevant columns to the cells containing the target answer, analogous to relational paths in graph settings. 
The supporting context set ($\mathcal{E}$) is constructed by aggregating triples that (i) involve question entities, (ii) contain the answer cell, or (iii) match semantically relevant column headers.

\begin{table}[t]
\centering
\caption{Cross-dataset quadrant analysis on \texttt{MetaQA-2hop} (Graph) and \texttt{WikiTableQuestions} (Table).}
\label{tab:combined_generalization}
\small
\setlength{\tabcolsep}{6pt}
\begin{tabular}{lccc}
\toprule
\multicolumn{4}{c}{\textbf{MetaQA-2hop (Graph)}} \\
\midrule
\textbf{Quadrant} & \textbf{Hall. (\%)} & \textbf{SSR} & \textbf{SAS} \\
\midrule
Q1: High SSR, High SAS & 36.36\% & 0.822 & 0.409 \\
Q2: Low SSR, High SAS  & \textbf{14.84\%} & 0.616 & 0.470 \\
Q3: Low SSR, Low SAS   & 18.39\% & 0.703 & 0.316 \\
Q4: High SSR, Low SAS  & 54.40\% & 0.821 & 0.321 \\
\midrule
\multicolumn{4}{c}{\textbf{WikiTableQuestions (Table)}} \\
\midrule
\textbf{Quadrant} & \textbf{Hall. (\%)} & \textbf{SSR} & \textbf{SAS} \\
\midrule
Q1: High SSR, High SAS & 84.11\% & 0.780 & 0.479 \\
Q2: Low SSR, High SAS  & \textbf{80.85\%} & 0.234 & 0.712 \\
Q3: Low SSR, Low SAS   & 87.54\% & 0.522 & 0.346 \\
Q4: High SSR, Low SAS  & 85.86\% & 0.777 & 0.354 \\
\bottomrule
\end{tabular}
\end{table}

\vspace{-0.2cm}

\begin{table*}
\centering
\setlength{\abovecaptionskip}{4pt}  
\caption{Performance comparison of hallucination detection methods across two LLMs. Bold denotes the best result and underline denotes the second-best.}
\label{tab:baseline_comparison}
\small 
\setlength{\tabcolsep}{3.5pt} 
\begin{tabular}{@{}lllccccc@{}}

\toprule
\textbf{LLM} & \textbf{Category} & \textbf{Method} & \textbf{AUC} 
& \textbf{\begin{tabular}[c]{@{}l@{}}F1\\ (class 1)\end{tabular}} 
& \textbf{\begin{tabular}[c]{@{}l@{}}F1\\ (macro avg)\end{tabular}} 
& \textbf{\begin{tabular}[c]{@{}l@{}}Precision\\ (class 1)\end{tabular}}
& \textbf{\begin{tabular}[c]{@{}l@{}}Recall\\ (class 1)\end{tabular}} \\
\midrule
\multirow{8}{*}{LLaMA2-7B}
& \multirow{3}{*}{Model-Internal Confidence}
& Perplexity                & 0.4578 & 0.2560 & 0.2126 &  0.1484 &  \textbf{0.9310} \\
&   & Token confidence          & 0.5020 & 0.2020 & 0.4529 & 0.1324 &  0.4249 \\
&   & Max token probability     & 0.5000 & \underline{0.2960} & 0.4478 & 0.1322 & 0.4404 \\
\addlinespace[3pt]
& \multirow{3}{*}{Semantic Consistency}
& BERTScore      & 0.6366 & 0.2740 & 0.5092 & 0.1820 & 0.5544 \\
&   & Embedding Divergence      & \underline{0.6914} & 0.2955 & \underline{0.5094} &  \underline{0.1914} & \underline{0.6476} \\
&   & NLI Contradiction      & 0.5741 & 0.2567 & 0.3524 &  0.1512 & 0.3667 \\
\addlinespace[3pt]
&  Mechanistic Detector(Ours)       & SSR + SAS  & \textbf{0.8341} & \textbf{0.5390} & \textbf{0.7524} & \textbf{0.5632} & 0.5168 \\
\midrule
\multirow{8}{*}{Qwen2.5-7B}
& \multirow{3}{*}{Model-Internal Confidence}
& Perplexity                & 0.4935  & 0.1086  & 0.3512  & 0.0602  & 0.5484 \\
&   & Token confidence          & 0.4966  & 0.1084  & \underline{0.3677}  & 0.0606  & 0.5161 \\
&   & Max token probability     & 0.4419  & \underline{0.1297}  & 0.1904  & \underline{0.0710}  & 0.7500 \\
\addlinespace[3pt]
& \multirow{2}{*}{Semantic Consistency}
& BERTScore       & 0.4867  & 0.1115  & 0.0745  & 0.0593 & \underline{0.9355} \\
&   & Embedding Divergence      & 0.4900  & 0.1134  & 0.0588  & 0.0602  & \textbf{0.9677} \\
&   & NLI Contradiction      & \underline{0.5044} & 0.1150 & 0.2625 &  0.0623 & 0.7419 \\
\addlinespace[3pt]
& Mechanistic Detector(Ours)         & SSR + SAS   & \textbf{0.8528} & \textbf{0.4606} & \textbf{0.7083} & \textbf{0.3974} & 0.5477 \\
\bottomrule
\end{tabular}
\end{table*}

\paragraph{Metric Discrimination and Correlation.}
Statistical analysis confirms that SSR and SAS remain effective discriminators of model reliability across datasets and knowledge formats. 
Truthful responses consistently exhibit significantly lower SSR and higher SAS than hallucinated ones (\(p < 0.001\)). 
Furthermore, the two metrics maintain a modest negative correlation (\(p < 0.001\)), confirming that they capture related yet distinct facets of the model's internal behavior. Detailed statistics and per-dataset results are provided in Appendix~\ref{generalization: stats}.

\paragraph{Joint Analysis.}
We further analyze hallucination behavior through joint SSR–SAS configurations (Table~\ref{tab:combined_generalization}).
Across all settings (1-hop, 2-hop, and tables), a consistent relationship emerges between Semantic Alignment Score (SAS) and model reliability, reflected in two stable cross-task patterns:

First, the quadrant characterized by the highest SAS (Q2) consistently exhibits the lowest hallucination rate in each setting, indicating that strong semantic grounding in the supporting context is reliably associated with faithful generation.
Second, the quadrants with the highest hallucination rates consistently fall within the low-SAS regime, regardless of whether the dominant failure mode manifests as diffuse attention (Q3) or shortcut-driven reasoning (Q4).

These findings align with the functional roles of different components in the Transformer architecture.
Structural reliance captured by SSR reflects the behavior of attention mechanisms, which govern how input tokens is selected and prioritized during reasoning. In contrast, semantic grounding captured by SAS corresponds to representation level alignment within FFNs. 
Notably, the representations used to compute SAS are extracted from the penultimate Transformer layer prior to output projection, whereas SSR is derived from attention patterns aggregated across decoder layers during generation.
Accordingly, while absolute hallucination risk varies with task complexity, SSR serves to modulate the specific modality of failure, whereas SAS provides a high-fidelity signal of whether the model will ultimately hallucinate, regardless of the underlying knowledge structure.

\subsection{From Mechanistic Signals to Lightweight Hallucination Detection} 
Building on these mechanistic signals, we demonstrate that SSR and SAS alone enable a lightweight, interpretable hallucination detector.


\paragraph{Detector Configuration}

We build a lightweight hallucination detector using the proposed mechanistic signals, SSR and SAS. 
For training and validation, we sample 37.5k instances from the \texttt{MetaQA-1hop} training set with an 80/20 split.
Using responses from two different base LLMs, \texttt{LLaMA2-7B} and \texttt{Qwen2.5-7B}, we train an XGBoost classifier on the resulting SSR and SAS signals.

\vspace{-0.075in}

\paragraph{Baselines and Results} 
We compare the proposed mechanistic hallucination detector against six widely adopted hallucination detection baselines, grouped into two categories:
(1) \emph{model-internal confidence signals}, including Perplexity \cite{manakul2023selfcheckgpt}, Token Confidence, and Max Token Probability ; and
(2) \emph{semantic consistency measures}, including BERTScore \cite{zhang2019bertscore}, Embedding Divergence \cite{reimers-gurevych-2019-sentence}, and NLI Contradiction \cite{bowman2015large}.

As shown in Table~\ref{tab:baseline_comparison}, the mechanistic hallucination detector consistently achieves the highest AUC and Macro-F1 scores across both LLMs (e.g., AUC $0.834$–$0.853$, F1 $0.461$-$0.540$, and Precision $0.397$–$0.563$).
While confidence-based and semantic similarity metrics often attain high recall, they suffer from substantially lower precision (e.g., Perplexity achieves a precision of $0.148$ on LLaMA2-7B), indicating a tendency to over-predict hallucinations.
In contrast, the joint use of SSR and SAS maintains a more balanced precision–recall trade-off, suggesting that integrating structural shortcut bias with representation-level semantic grounding offers a more reliable indicator of hallucination than raw probability or surface-level semantic similarity alone.

Beyond performance, the proposed signals offer advantages in efficiency and interpretability.
SSR and SAS are computed within a single model inference, avoiding the overhead of sampling-based or self-consistency methods. 
Furthermore, since SSR and SAS correspond to internal failure mechanisms, detection outcomes admit a natural interpretation in terms of underlying model behavior. Detailed ablation studies confirming the complementary nature of these two signals are provided in Appendix~\ref{sec: detector_ablation}.

\section{Conclusion}
This work analyzes hallucination in large language models when reasoning over structured knowledge that has been linearized into token sequences. We introduce two mechanistic signals, Structural Shortcut Reliance (SSR) and Semantic Alignment Score (SAS), to characterize complementary internal failure mechanisms in attention and feed forward representations. Across graph and table reasoning tasks, we find that while attention patterns vary with task structure, failures in semantic grounding within feed forward layers are consistently associated with hallucination. 
Building on these insights, we further demonstrate the generalizability of our findings and develop a lightweight and interpretable hallucination detector based on the proposed mechanistic signals. Without requiring any model fine-tuning, this detector consistently outperforms standard confidence-based and similarity-based baselines. Our results highlight the central role of representation-level semantic grounding in enabling reliable reasoning over structured knowledge. We believe that the proposed framework is broadly reusable for future research on hallucination analysis and mitigation. Our code and implementation are available at: \href{https://github.com/ShanghaoLi0913/struhall-mechanism}{https://github.com/ShanghaoLi0913/struhall-mechanism}.

\section{Limitations}
While our analysis provides insights into hallucination in structured reasoning, several limitations remain.
First, we focus on decoder-only Transformers; other architectures may exhibit different internal dynamics. 
Second, we study linearized structured knowledge, and do not explore alternative representations such as graph-aware encodings or specialized prompting strategies. 
Third, our analysis is observational, capturing correlations between internal signals and hallucination; intervention-based studies could further clarify semantic grounding dynamics. 
Finally, due to computational constraints, we evaluate models up to 7B parameters, leaving behavior in larger models (e.g., 70B+) for future investigation.

\section{Ethical Considerations}
This work enhances the transparency of LLM-based reasoning in knowledge-intensive tasks by improving the interpretability of hallucination behaviors, thereby contributing to the reduction of misinformation in real-world deployments. While our proposed signals provide valuable diagnostic insights, they are intended as analytical tools rather than standalone safety solutions. We caution against over-reliance on automated detectors in high-stakes domains and emphasize that these metrics should complement, rather than replace, rigorous human oversight. To facilitate responsible use and further research, all experiments utilize publicly available datasets and open-source models, and we commit to releasing our codebase and evaluation scripts upon acceptance.

\section*{Acknowledgments}
This work is supported in part by NSF under grants III-2106758, and POSE-2346158.

\bibliography{custom}
\appendix
\section{Subgraph Construction and Pruning Procedure}
\label{appendix:pruning}
This appendix details the pipeline for transforming raw knowledge graph (KG) data into pruned subgraphs suitable for LLM-based question answering. The procedure is designed to balance the inclusion of critical reasoning paths with a manageable input size that preserves semantic context.

\subsection{Core Structural Cues Extraction}
The extraction of \emph{Core Structural Cues} ($S$) begins with identifying question entities through exact word matching against the entity vocabulary, ensuring matches respect whole-word boundaries. We then represent the KG as an undirected graph $G = (V, E)$ to facilitate pathfinding. For every pair of question entities $q \in Q$ and gold answer entities $a \in A$, we execute a breadth-first search (BFS) to discover all shortest paths $P_{q \to a}$ within a maximum depth of 3 hops. All triples $(e_i, r, e_{i+1})$ residing along these discovered paths are collected to form the set $S$, representing the minimal skeletal structure required to reach the answer.

\subsection{Subgraph Trimming and Importance Scoring}
To construct a manageable subgraph while maintaining answer reachability, we implement a shortest-path-plus-neighbors strategy. We initialize the subgraph with all entities and triples from $S$ (denoted as $E_{path}$) and then expand the set by identifying 1-hop neighbor triples connected to any $e \in E_{path}$. To prevent context overflow, each neighbor triple $t$ is prioritized using a connectivity-based heuristic:
\begin{equation}
\begin{split}
\text{score}(t) = & \, 3 \cdot \mathds{1}[t \cap E_{\text{path}} \neq \emptyset] \\
& + 2 \cdot \mathds{1}[t \cap Q \neq \emptyset] + 2 \cdot \mathds{1}[t \cap A \neq \emptyset]
\end{split}
\end{equation}
Triples are sorted by this score, and the top-scoring candidates are selected until the subgraph reaches a size of $K=20$ (or a dataset-specific threshold). Finally, we validate the subgraph to guarantee that at least one gold answer entity remains reachable, ensuring the task is theoretically solvable from the provided context.

\subsection{Supporting Context Set Construction}
The \emph{Supporting Context Set} ($\mathcal{E}$) serves to provide the semantic ``meat'' around the skeletal cues in $S$ for alignment analysis. We define a core entity set $C$ containing all entities present in $S$. We then scan the trimmed subgraph for neighbor triples $(h, r, t)$ where at least one participant belongs to $C$. To ensure the context is informative, we apply a semantic filter that retains only triples whose relations match predefined patterns indicative of domain knowledge (e.g., \texttt{directed\_by}, \texttt{genre}). 

To prevent highly connected ``hub'' entities from dominating the context, we enforce a per-entity expansion limit of 3 triples. The final set $\mathcal{E}$ is formed by the union of $S$ and these neighbors, truncated at a total of 20 triples to fit within model context windows while prioritizing the retention of $S$.

\begin{figure}[h]
   \centering
   \setlength{\abovecaptionskip}{3pt}
   \setlength{\belowcaptionskip}{-3pt}
   \includegraphics[width=\linewidth]{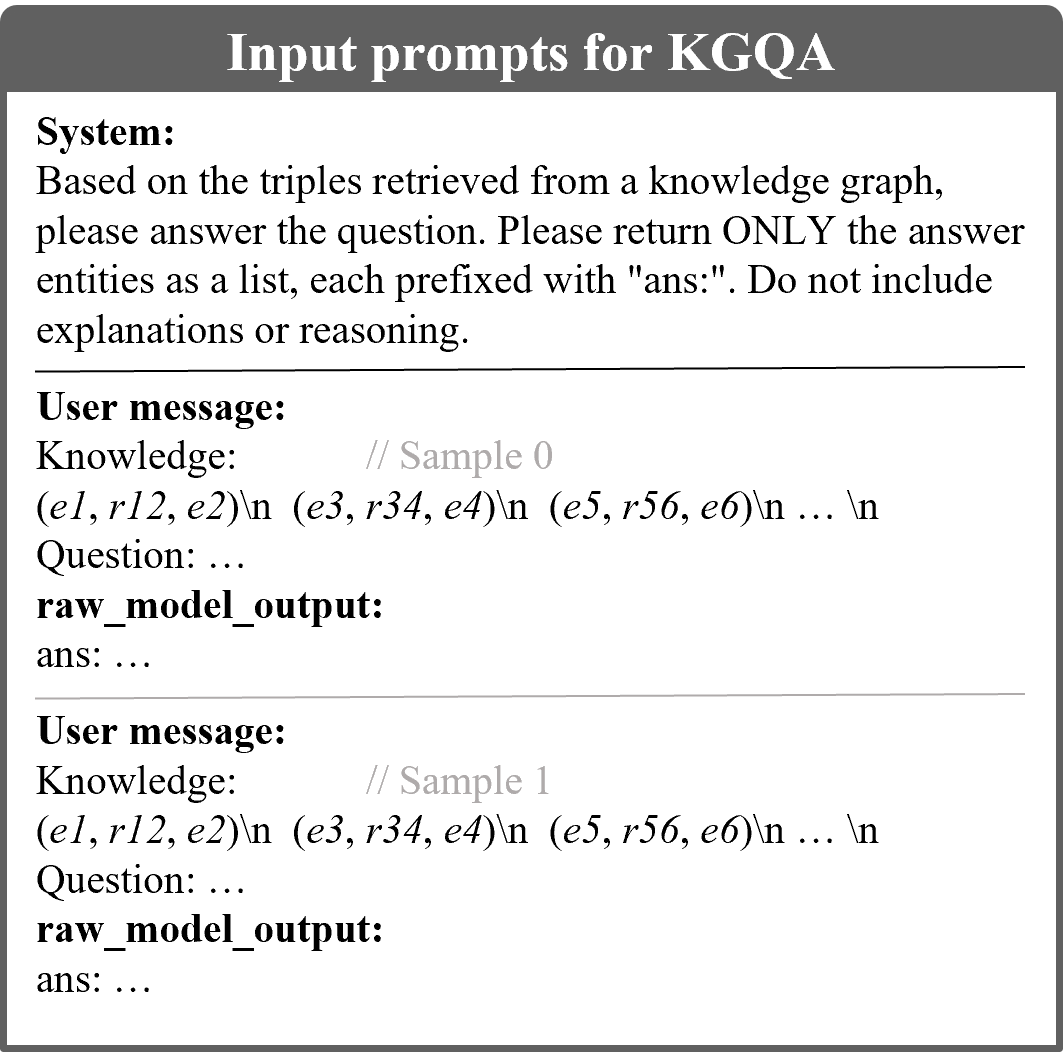}
   \caption{Prompt template for graph-based question answering, illustrating linearized knowledge graph triples and constrained answer format.}
   \label{fig:prompt_template_figure}
\end{figure}

\subsection{Linearization and Prompting}
Structured knowledge is serialized into a natural language format by converting each triple $(h, r, t)$ into a simple string $h$ $r$ $t$'' using textual labels. These serialized triples are then injected into a standardized prompt template. As illustrated in Figure~\ref{fig:prompt_template_figure}, the prompt provides the reasoning context followed by the question, instructing the model to output answer entities in a constrained format prefixed with ``ans:'' for consistent evaluation.

\section{Interpretation of Mechanistic Metrics}
\label{sec:metric_interpretation}
\paragraph{Interpretation of Structural Shortcut Reliance (SSR).}
A \emph{high} SSR indicates that the decoder consistently prioritizes core structural cues over supporting context, reflecting an over-reliance on structural shortcuts. 
Conversely, a \emph{low} or negative SSR signifies more distributed attention allocation, suggesting that the model actively integrates broader evidence, which correlates with higher factual accuracy.

\paragraph{Interpretation of Semantic Alignment Score (SAS).}
A \emph{high} SAS indicates that the feed-forward networks (FFNs) successfully incorporate the source knowledge provided in the input into the model’s internal representations during generation. 
In contrast, a \emph{low} SAS suggests representation drift, where the internal state becomes weakly coupled with the input context and instead reflects pretrained parametric associations, increasing the risk of hallucination.

\section{Additional Robustness Analyses}
\label{sec:robustness}

We provide additional evidence to assess the robustness of our findings across model scales, reasoning depths, and variations in input structure. These analyses are intended to verify that the observed SSR–SAS patterns are not artifacts of specific datasets, models, or linearization choices.

\subsection{Robustness Across Model Scales}

To examine whether the observed patterns persist on newer model families, we evaluate \texttt{Qwen3-8B} and \texttt{Qwen3-14B} under the same experimental setup.

Across both models, we observe consistent distributional separation between truthful and hallucinated outputs: hallucinated samples exhibit higher SSR and lower SAS, with statistically significant differences ($p < 0.001$). The SSR–SAS quadrant structure also remains qualitatively stable, with the low-SAS regimes consistently associated with higher hallucination rates.

These results suggest that the relationship between structural shortcutting and semantic alignment is not restricted to earlier model families, but persists across more recent architectures within the evaluated scale range.

\subsection{Robustness Across Reasoning Depths}
We further evaluate the proposed metrics on more complex multi-hop reasoning settings (see Table~\ref{tab:depth}).

On \texttt{MetaQA-3hop}, the same directional trends persist: hallucinated outputs exhibit higher SSR and lower SAS than truthful ones ($p < 0.001$). The quadrant structure remains consistent, with the low-SSR + low-SAS region exhibiting the highest hallucination rate.

On \texttt{ComplexWebQuestions} (CWQ, up to 4-hop), despite increased structural complexity, we observe the same pattern: hallucinated samples show higher SSR and lower SAS relative to truthful outputs, with statistically significant differences ($p < 0.01$ for SSR and $p < 0.001$ for SAS). The relative ordering of SSR--SAS quadrants remains consistent with the 1-hop setting, with the low-SSR, high-SAS region consistently exhibiting the lowest hallucination rate.

These findings indicate that the interaction between structural shortcuts and semantic misalignment persists under deeper reasoning chains and more complex query structures.

\begin{table}[h]
\centering
\small
\caption{Robustness across reasoning depths.}
\label{tab:depth}
\begin{tabular}{l c c c}
\toprule
\textbf{Dataset} & \textbf{SSR (H)} & \textbf{SSR (T)} & \textbf{$p$-value} \\
\midrule
MetaQA (3-hop) & 0.7339 & 0.6424 & $< 0.001$ \\
CWQ (4-hop) & 0.5947 & 0.5233 & $< 0.01$ \\
\bottomrule
\end{tabular}
\end{table}

\subsection{Sensitivity to Linearization Order}
\label{sec: linearization order}
To assess whether SSR is driven by sequence order artifacts, we randomly permute the order of non-supporting triples in the linearized input while keeping the supporting triples (i.e., those on the shortest path connecting question entities to the answer) fixed in their original positions, and recompute SSR.

The resulting signal remains highly stable, with a Spearman correlation of $\rho = 0.881$ ($p < 0.001$) between original and permuted configurations. This suggests that SSR is not primarily determined by positional biases in the linearized sequence, but reflects a more stable attention distribution pattern over supporting structures.

\subsection{Sensitivity to Support-Set Definition}
\label{sec: support set}
We further evaluate the robustness of SSR under variations in the definition of the supporting context. Specifically, we compare (1) a minimal support set consisting of the shortest-path triples connecting the question entities to the answer, and (2) an expanded support set that augments these triples with additional neighboring facts retrieved from the local graph structure. 

Across two variants, the directional separation between truthful and hallucinated samples remains consistent. The discriminative behavior of SSR is also stable, with comparable performance across settings ($mean AUC = 0.562, std = 0.010$).

These results suggest that SSR captures a general tendency toward structural shortcutting associated with minimal supporting structures, rather than being tied to a specific heuristic instantiation.

\section{Detailed Statistical Results}
\label{generalization: stats}
As shown in table~\ref{tab:detailed_stats_appendix}, this appendix provides the detailed statistical results supporting the generalization analysis in Section~6.1.

\begin{table}
    \centering
    \caption{Detailed statistics of SSR and SAS for truthful and hallucinated samples.
    All differences are statistically significant ($p < 0.001$).}
    \label{tab:detailed_stats_appendix}
    \small
    \begin{tabular}{l c c c c}
        \toprule
        \textbf{Category} & \textbf{SSR} & \textbf{SAS} & \textbf{$t_{SSR}$} & \textbf{$t_{SAS}$} \\
        \midrule
        \multicolumn{5}{l}{\textbf{WikiTableQuestions}} \\
        \midrule
        Truthful     & 0.493 & 0.537 & \multirow{2}{*}{$-3.91$} & \multirow{2}{*}{$4.35$} \\
        Hallucinated & 0.563 & 0.486 &  &  \\
        \midrule
        \multicolumn{5}{l}{\textbf{MetaQA-2hop (Dev)}} \\
        \midrule
        Truthful     & 0.701 & 0.395 & \multirow{2}{*}{$-8.60$} & \multirow{2}{*}{$6.28$} \\
        Hallucinated & 0.793 & 0.342 &  &  \\
        \bottomrule
    \end{tabular}
\end{table}

\noindent\textbf{Correlation Analysis:} 
SSR and SAS maintain a modest negative correlation across both domains: \(r \approx -0.18\) for WikiTableQuestions and \(r \approx -0.22\) for MetaQA-2hop (both \(p < 0.001\)). This indicates that structural shortcut reliance and semantic misalignment are systematically related yet sufficiently decoupled to reflect distinct internal aspects of model behavior.

\vspace{-0.2\baselineskip} 
\begin{table}
\centering
\setlength{\abovecaptionskip}{2pt} 
\setlength{\belowcaptionskip}{-3pt} 
\caption{Ablation results across two different metrics. The boldface represents the best performance, and the underline represents the second-best.}
\label{tab:ablation}
\small
\begin{tabular}{@{}l|cc|cc@{}}
\toprule
\textbf{Feature Set} & \multicolumn{2}{c|}{\textbf{LLaMA2-7B}} & \multicolumn{2}{c}{\textbf{Qwen2.5-7B}} \\
 & AUC & Recall & AUC & Recall \\
\midrule
SSR-only   & 0.5502 & 0.4763 & 0.5067 & \textbf{0.6770} \\
SAS-only   & \underline{0.6324} & \textbf{0.7291} &  \underline{0.6331} & \underline{0.6527} \\
\textbf{SSR + SAS} & \textbf{0.8328} & \underline{0.5223} & \textbf{0.8506} & 0.5300 \\
\bottomrule
\end{tabular}
\end{table}

\section{Ablation Study on Mechanistic Signals}
\label{sec: detector_ablation}
We investigate the individual contributions of Structural Shortcut Reliance (SSR) and Semantic Alignment Score (SAS) to the overall performance of our mechanistic detector. 

As shown in Table~\ref{tab:ablation}, while SAS-only and SSR-only can each achieve high recall (up to 0.729 and 0.677, respectively), their AUC values remain moderate (0.550--0.633), indicating limited overall discriminative power when used alone in detecting hallucinations.
In contrast, integrating both metrics (SSR + SAS) leads to a substantial increase in AUC, exceeding 0.833 across models.

This result highlights the complementarity between attention-level structural shortcut bias and FFN-level semantic grounding failure, yielding more reliable hallucination identification by enhancing discriminative power rather than relying on recall alone.

\end{document}